\pdfoutput=1

\documentclass[11pt]{article}

\usepackage{acl}

\usepackage{times}
\usepackage{latexsym}
\usepackage{graphicx}
\usepackage{float}

\usepackage{hyperref}
\usepackage{lipsum}
\usepackage{booktabs}
\usepackage{placeins}
\usepackage{algorithm2e}
\usepackage{amsmath}
\usepackage{amssymb}
\usepackage{multirow}
\usepackage{subcaption}
\usepackage{url}            
\usepackage{nicefrac}       
\usepackage{microtype}      
\usepackage{enumitem}
\usepackage{listings}
\usepackage{tcolorbox}
\usepackage{xspace}
\usepackage{soul}               
\setstcolor{red}            
\usepackage{tikz}
\usepackage{xcolor} 

\usepackage[T1]{fontenc}

\usepackage[utf8]{inputenc}

\usepackage{microtype}

\usepackage[draft,textsize=footnotesize,textwidth=15mm]{todonotes}

\definecolor{paired-light-blue}{RGB}{198, 219, 239}
\definecolor{paired-dark-blue}{RGB}{49, 130, 188}
\definecolor{paired-light-orange}{RGB}{251, 208, 162}
\definecolor{paired-dark-orange}{RGB}{230, 85, 12}
\definecolor{paired-light-green}{RGB}{199, 233, 193}
\definecolor{paired-dark-green}{RGB}{49, 163, 83}
\definecolor{paired-light-purple}{RGB}{218, 218, 235}
\definecolor{paired-dark-purple}{RGB}{117, 107, 176}
\definecolor{paired-light-gray}{RGB}{217, 217, 217}
\definecolor{paired-dark-gray}{RGB}{99, 99, 99}
\definecolor{paired-light-pink}{RGB}{222, 158, 214}
\definecolor{paired-dark-pink}{RGB}{123, 65, 115}
\definecolor{paired-light-red}{RGB}{231, 150, 156}
\definecolor{paired-dark-red}{RGB}{131, 60, 56}
\definecolor{paired-light-yellow}{RGB}{231, 204, 149}
\definecolor{paired-dark-yellow}{RGB}{141, 109, 49}
\definecolor{Dandelion}{HTML}{FF9966}

\definecolor{bg1}{HTML}{FF9966}
\definecolor{bg2}{HTML}{CCE5FF}
\definecolor{bg3}{HTML}{FFCC99}
\definecolor{bg4}{HTML}{FFC107}
\definecolor{bg5}{HTML}{FFCCCC}
\definecolor{bg6}{HTML}{D5E8D4}
\definecolor{bg7}{HTML}{eeeeee}
\definecolor{bg8}{HTML}{cdeb8b}
\definecolor{bg9}{HTML}{dae8fc}
\definecolor{bg10}{HTML}{a2e6eb}

\definecolor{bg31}{HTML}{FFCDD2} 

\definecolor{bg32}{HTML}{F8BBD0}

\definecolor{bg33}{HTML}{E1BEE7} 

\definecolor{bg34}{HTML}{D7CCC8} 

\definecolor{bg35}{HTML}{B2DFDB} 

\definecolor{bg36}{HTML}{A5D6A7} 

\definecolor{bg37}{HTML}{FFF9C4} 

\definecolor{bg38}{HTML}{FFECB3} 

\definecolor{bg111}{HTML}{CB6843}

\definecolor{bg112}{HTML}{D77C5C}

\definecolor{bg113}{HTML}{E28E6E}
\definecolor{bg114}{HTML}{E89F7D}
\definecolor{bg115}{HTML}{EDAE8A}
\definecolor{bg116}{HTML}{F0BA95}
\definecolor{bg117}{HTML}{F3C29F}
\definecolor{bg118}{HTML}{F6CCAA}
\definecolor{bg119}{HTML}{F8D5B3}
\definecolor{bg120}{HTML}{FADCBD}
\definecolor{bg121}{HTML}{FCE6C7}

\definecolor{bg39}{HTML}{FFE0B2} 

\definecolor{bg40}{HTML}{3CB371} 

\definecolor{bg43}{HTML}{ffe5d9}

\definecolor{bg15}{HTML}{7FFFD4}

\definecolor{bg17}{HTML}{F0FFFF}

\definecolor{bg18}{HTML}{F5FFFA}

\definecolor{bg19}{HTML}{F8F8FF}

\definecolor{bg20}{HTML}{FFFFFF}

\definecolor{bg21}{HTML}{E1F5FE}

\definecolor{bg22}{HTML}{B3E5FC}

\definecolor{bg23}{HTML}{81D4FA}

\definecolor{bg24}{HTML}{4FC3F7}

\definecolor{bg25}{HTML}{29B6F6}

\definecolor{bg26}{HTML}{03A9F4}

\definecolor{bg27}{HTML}{039BE5}

\definecolor{bg28}{HTML}{0288D1}

\definecolor{bg29}{HTML}{0277BD}

\definecolor{bg30}{HTML}{01579B}

\definecolor{bg16}{HTML}{FFCC99}

\definecolor{pg51}{HTML}{33cccc} 
\definecolor{pg52}{HTML}{C8E6C9} 
\definecolor{pg53}{HTML}{B9F6CA} 
\definecolor{pg54}{HTML}{A9DFBF} 
\definecolor{pg55}{HTML}{BCF5A6} 

\definecolor{pg56}{HTML}{BEF1CE} 
\definecolor{pg57}{HTML}{CEF6EC} 
\definecolor{pg58}{HTML}{32de84} 
\definecolor{pg59}{HTML}{B1F2B5} 
\definecolor{pg60}{HTML}{FFD700} 

\definecolor{pg61}{HTML}{DEF7E0} 
\definecolor{pg62}{HTML}{E8F8DC} 

\definecolor{pg63}{HTML}{EBF7E7} 
\definecolor{pg64}{HTML}{F0FDF4} 

\definecolor{pg65}{HTML}{F1FEE7} 
\definecolor{pg66}{HTML}{F7FFF6} 
\definecolor{pg67}{HTML}{FCFFE7} 
\definecolor{pg68}{HTML}{D8BFD8} 
\definecolor{pg69}{HTML}{EEFFE2} 
\definecolor{pg70}{HTML}{6CB4EE} 

\definecolor{connect-color}{RGB}{0,0,0}
\definecolor{middle-color}{RGB}{255,255,255}
\definecolor{leaf-color}{RGB}{173,216,230}
\definecolor{line-color}{RGB}{25,25,112}

\usepackage{longtable}
\usepackage{tablefootnote}
\usepackage[edges]{forest}
\definecolor{hidden-draw}{RGB}{20,68,106}
\definecolor{hidden-pink}{RGB}{255,245,247}
\definecolor{red}{RGB}{255,0,0}

\definecolor{hidden-draw}{RGB}{0,0,0}
\definecolor{hidden-pink}{RGB}{255,182,193}

\title{A Comprehensive Survey of Hallucination in Large Language, Image, Video and Audio Foundation Models}

\author{
    Pranab Sahoo$^1$,
    Prabhash Meharia$^1$,
    Akash Ghosh$^1$,
    Sriparna Saha$^1$, \\
    Vinija Jain$^{2}$,
    and Aman Chadha$^{2,3}$\thanks{\,\,\,Work does not relate to position at Amazon.}\\
    \vspace{0.2cm}
    $^1$Department of Computer Science and Engineering, Indian Institute of Technology Patna\\
    $^2$Stanford University, $^3$Amazon GenAI\\
    \vspace{0.1cm}
    \small
    \begin{tabular}[t]{@{}c@{}}
        \texttt{pranab\_2021cs25@iitp.ac.in, prabhash\_2211cs12@iitp.ac.in, sriparna@iitp.ac.in}\\
        \texttt{hi@vinija.ai, hi@aman.ai}
    \end{tabular}
}

\begin{document}
\maketitle
\begin{abstract}
The rapid advancement of foundation models (FMs) across language, image, audio, and video domains has shown remarkable capabilities in diverse tasks. However, the proliferation of FMs brings forth a critical challenge: the potential to generate hallucinated outputs, particularly in high-stakes applications. The tendency of foundation models to produce hallucinated content arguably represents the biggest hindrance to their widespread adoption in real-world scenarios, especially in domains where reliability and accuracy are paramount. This survey paper presents a comprehensive overview of recent developments that aim to identify and mitigate the problem of hallucination in FMs, spanning text, image, video, and audio modalities. By synthesizing recent advancements in detecting and mitigating hallucination across various modalities, the paper aims to provide valuable insights for researchers, developers, and practitioners. Essentially, it establishes a clear framework encompassing definition, taxonomy, and detection strategies for addressing hallucination in multimodal foundation models, laying the foundation for future research and development in this pivotal area.
\end{abstract}

\section{Introduction}
The rapid progress in large-scale foundation models (FMs), spanning language, image, audio, and video domains, has revolutionized the field of artificial intelligence (AI). Models such as GPT-3~\cite{brown2020language}, MiniGPT-4~\cite{zhu2023minigpt4}, AudioLLM~\cite{borsos2023audiolm}, and LaViLa~\cite{zhao2022learning} have demonstrated remarkable abilities across diverse tasks, from text generation to multimodal understanding. As these models find wider applications in critical domains, there is a growing imperative to comprehend and alleviate their propensity to produce hallucinated outputs. 

\subsection{Hallucination} 
Hallucination refers to instances where FMs generate plausible but factually incoherent or absurd content lacking proper context understanding~\cite{xu2024hallucination}. These hallucinated outputs can range from minor inaccuracies to completely imaginary content, manifesting across text, images, videos, and audio generated by large models. Root causes may include biases in training data, limited access to up-to-date information, or inherent model constraints in contextual comprehension and response generation. Deploying such powerful models without addressing hallucinations risks perpetuating misinformation, incorrect conclusions, and adverse effects in critical applications. Mitigating hallucinations has become an active research focus, with strategies like fine-tuning with domain-specific data, using diverse, robust training data, and developing improved evaluation metrics to identify and reduce hallucination tendencies.


\subsection{Types of Hallucination}
\begin{figure} 
 \centering
\includegraphics[width=\linewidth]{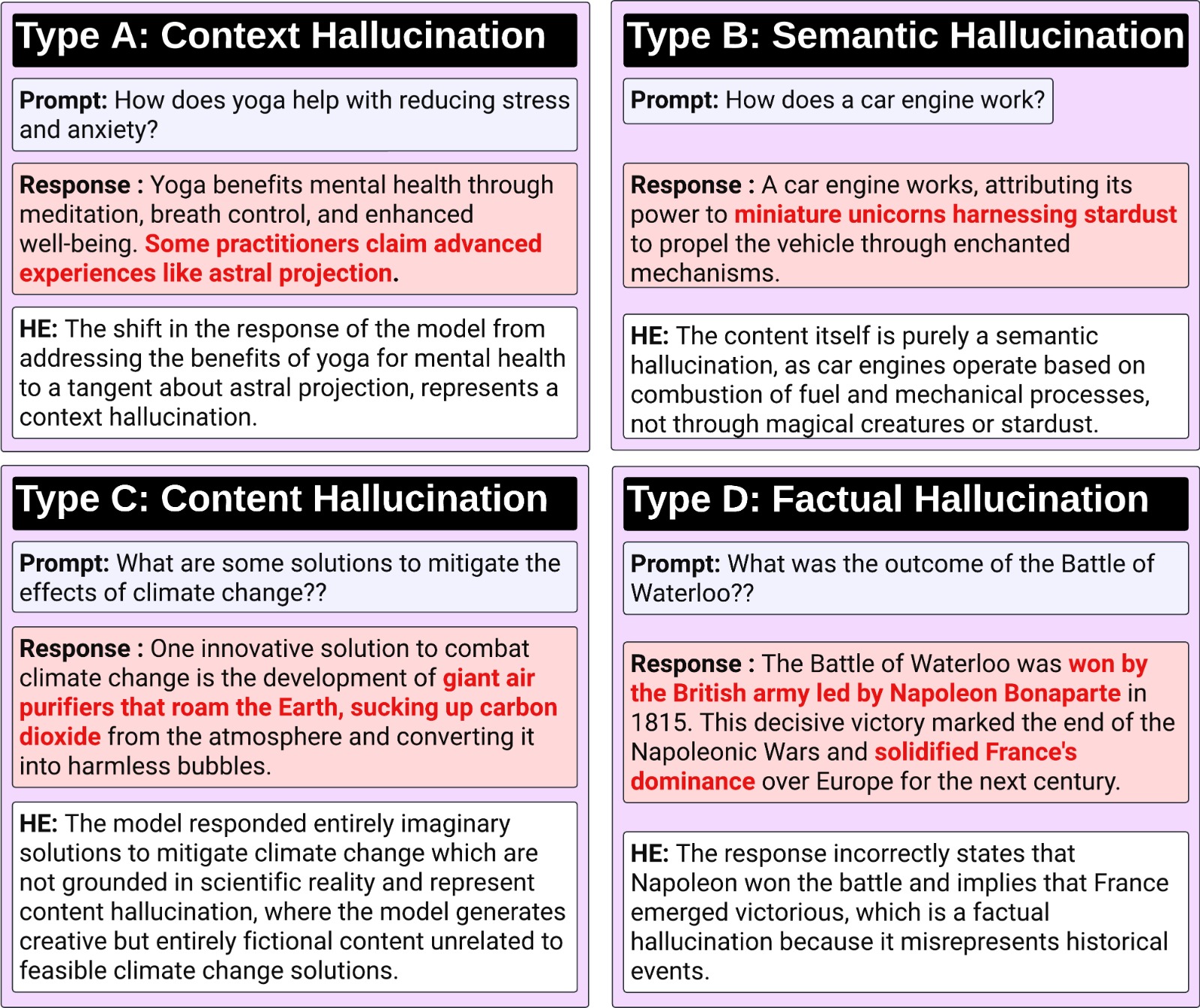}
 \caption{Illustration of Hallucination types. Proper explanations of hallucinations are indicated as hallucinated elements (HE) and are highlighted in bold red text.}
\label{typeHallu}
\end{figure}

Hallucinations in large FMs can manifest in various forms and can be categorized as follows:
\textbf{Contextual disconnection:}~\citet{zhang2023siren}~described a situation in which the output or content produced by a model across different modalities is inconsistent or out of sync with the context that the user or the input data provided or expected. \textbf{Semantic distortion:}~ \citet{tjio2022adversarial} refers to a type of inconsistency or error in generated content where the semantics or underlying meaning of the input is misrepresented or altered in the output. \textbf{Content hallucination} is the term used to describe a phenomenon seen in generative models when features or elements that are generated as output are either unreal given the context or absent from the input data~\citet{moernaut2018content}. \textbf{Factual inaccuracy:}~ \citet{zhang2023siren} described a kind of error seen in generative models when information that is inaccurate, deceptive, or at odds with the known facts appears in the generated output. Figure~\ref{typeHallu} illustrates various types of hallucinations with examples.

\subsection{Motivation and Contributions}
Most of the existing survey papers have explored hallucination in the context of large language models (LLMs)~\cite{huang2023survey},~\cite{chen2024unified},~\cite{tonmoy2024comprehensive}. Recent studies have shown that hallucination can also occur in vision, audio, and video foundation models, highlighting the need for a comprehensive understanding of this challenge across multiple modalities~\cite{liu2024survey},~\cite{sahoo2024systematic},~\cite{rawte2023survey}. To address this gap, the present survey aims to provide a holistic and multimodal perspective on the hallucination challenge in FMs across language, vision, video, and audio domains. It serves as a vital resource for researchers and practitioners, aiding in the development of more robust AI solutions. Additionally, it includes a detailed taxonomy diagram in Fig.~\ref{taxo} and a summarized Table~\ref{R1} (refer to section~\ref{ta} of the appendix) illustrating recent advancements across different modalities. The contributions of this survey paper are as follows:


\begin{itemize}
     \item This survey provides a comprehensive overview of hallucination detection and mitigation techniques designed specifically for multimodal foundation models, extending beyond the typical focus solely on language models.
    \item Establish a precise definition and structured taxonomy of hallucination in the context of large-scale foundation models. 
    \item We have presented the various detection and mitigation strategies that have been proposed to address the hallucination problem in a multimodal setting.
    \item Highlight the open challenges and future research directions in this critical area.
\end{itemize}

\begin{figure*}[ht!]
  \centering
  \resizebox{1\textwidth}{!}{%
    \begin{forest}
      forked edges,
      for tree={
        grow=east,
        reversed=true,
        anchor=base west,
        parent anchor=east,
        child anchor=west,
        base=center,
        font=\large,
        rectangle,
        draw=hidden-draw,
        rounded corners,
        align=center,
        text centered,
        minimum width=5em,
        edge+={darkgray, line width=1pt},
        s sep=3pt,
        inner xsep=2pt,
        inner ysep=3pt,
        line width=0.8pt,
        ver/.style={rotate=90, child anchor=north, parent anchor=south, anchor=center},
      },
      where level=1{text width=15em,font=\normalsize,}{},
      where level=2{text width=14em,font=\normalsize,}{},
      where level=3{minimum width=10em,font=\normalsize,}{},
      where level=4{text width=26em,font=\normalsize,}{},
      where level=5{text width=20em,font=\normalsize,}{},
      [
        \textbf{Hallucinations}, for tree={fill=paired-light-red!80}, text width=14em
        [
            \textbf{Text} \S\ref{LLM}, for tree={fill=red!50}, text width=18em
              [
                    \textbf{Detection}\\
                     [
                    \textbf{Automatic Evaluation}
                      [
                        \textbf{FACTOID} \cite{rawte2024factoid}, for tree={fill=pg56},text width=28em
                      ]
                      [
                        \textbf{FACTOR} \cite{muhlgay2023generating}, for tree={fill=pg56},text width=28em
                      ]
                      [
                        \textbf{FactCHD} \cite{chen2023unveiling}, for tree={fill=pg56},text width=28em
                      ]
                  ]
                  [
                    \textbf{Inference Classifier}
                      [
                        \textbf{HalluQA} \cite{cheng2023evaluating}, for tree={fill=pg56},text width=28em
                      ]
                      [
                        \textbf{HaluEval} \cite{li2023halueval}, for tree={fill=pg56},text width=28em
                      ]
                  ]
                  [
                    \textbf{Self-Evaluation}
                       [
                        \textbf{SelfCheckGPT} \cite{manakul2023selfcheckgpt}, for tree={fill=pg56},text width=28em
                      ]
                  ]
                  [
                    \textbf{Evidence Retrieval}
                      [
                        \textbf{FACTSCORE} \cite{min2023factscore}, for tree={fill=pg56},text width=28em
                      ]
                      [
                        \textbf{FacTool} \cite{chern2023factool}, for tree={fill=pg56},text width=28em
                      ]
                  ]
              ]
              [
                    \textbf{Mitigation}\\
                    [
                    \textbf{Data Augmentation and Manipulation}
                       [
                        \textbf{DoLa} \cite{chuang2023dola}, for tree={fill=pg58},text width=28em
                      ]
                      [
                            \textbf{LLM-AUGMENTER} \cite{peng2023check}, for tree={fill=pg58},text width=28em
                      ]
                  ]
                  [
                      \textbf{Chain-of-Thought and Reasoning Enhancement}
                      [
                        \textbf{CoK} \cite{li2023chain}, for tree={fill=pg58},text width=28em
                      ]
                      [
                        \textbf{CoVe} \cite{dhuliawala2023chain}, for tree={fill=pg58},text width=28em
                      ]
                  ]
                  [
                    \textbf{Alignment and Fine-Tuning}
                      [
                        \textbf{MixAlign} \cite{knowledgeknowledge}, for tree={fill=pg58},text width=28em
                      ]
                      [
                        \textbf{PURR} \cite{chen2023purr}, for tree={fill=pg58},text width=28em
                      ]
                  ]
                  [
                    \textbf{Self-Verification and Consistency Checking}
                      [
                        \textbf{SELF-FAMILIARITY} \cite{luo2023zero}, for tree={fill=pg58},text width=28em
                      ]
                      [
                        \textbf{HALOCHECK} \cite{elaraby2023halo}, for tree={fill=pg58},text width=28em
                      ]
                  ]
                  [
                    \textbf{Prompt Engineering and Instructional Techniques}
                      [
                        \textbf{Instructional Prompting} \cite{varshney2023stitch}, for tree={fill=pg58},text width=28em
                      ]
                  ]  
              ]
        ]
        [
            \textbf{Image} \S\ref{VLM1}, for tree={fill=orange!50}, text width=18em
             [
                    \textbf{Detection}\\
                  [
                    \textbf{Human Evaluation and Cognitive Models}\\
                      [
                        \textbf{GAVIE} \cite{liu2023mitigating}, for tree={fill=Dandelion},text width=28em
                      ]
                      [
                        \textbf{PhD} \cite{liu2024phd}, for tree={fill=Dandelion},text width=28em
                      ]
                      [
                        \textbf{REVO-LION} \cite{liao2023revo}, for tree={fill=Dandelion},text width=28em
                      ]
                  ]
                  [
                    \textbf{Visual Question Answering}\\
                      [
                        \textbf{VHTest} \cite{huang2024visual}, for tree={fill=Dandelion},text width=28em
                      ]
                      [
                        \textbf{VQA} \cite{changpinyo2022maxm}, for tree={fill=Dandelion},text width=28em
                      ]
                  ]
                  [
                    \textbf{Benchmark and Evaluation Datasets}\\
                      [
                        \textbf{HallusionBench} \cite{guan2023hallusionbench}, for tree={fill=Dandelion},text width=28em
                      ]
                      [
                        \textbf{ChartBench} \cite{xu2023chartbench}, for tree={fill=Dandelion},text width=28em
                      ]
                      [
                        \textbf{MMHAL-BENCH} \cite{sun2023aligning}, for tree={fill=Dandelion},text width=28em
                      ]
                      [
                        \textbf{TouchStone} \cite{bai2023touchstone}, for tree={fill=Dandelion},text width=28em
                      ]
                  ]
                  [
                    \textbf{Multimodal Evaluation}\\
                      [
                        \textbf{MMVP} \cite{tong2024eyes}, for tree={fill=Dandelion},text width=28em
                      ]
                      [
                        \textbf{M-HalDetect} \cite{gunjal2024detecting}, for tree={fill=Dandelion},text width=28em
                      ]
                  ]
                  [
                    \textbf{Quantitative Metrics and Scoring Systems}\\
                      [
                        \textbf{FAITHSCORE} \cite{jing2023faithscore}, for tree={fill=Dandelion},text width=28em
                      ]
                      [
                        \textbf{POPE} \cite{li2023evaluating}, for tree={fill=Dandelion},text width=28em
                      ]
                      [
                        \textbf{NOPE} \cite{lovenia2023negative}, for tree={fill=Dandelion},text width=28em
                      ]
                   ]
                   [
                    \textbf{CIEM} \cite{hu2023ciem}, for tree={fill=Dandelion},text width=28em
                  ]
                  [
                    \textbf{ MSG-MCQ} \cite{lu2024evaluation}, for tree={fill=Dandelion},text width=28em
                  ]
                  [
                    \textbf{HaELM} \cite{wang2023evaluation}, for tree={fill=Dandelion},text width=28em
                  ]
              ]
              [
                    \textbf{Mitigation}\\
                    [
                      \textbf{Context and Semantic Understanding}\\
                      [
                        \textbf{HalluciDoctor} \cite{yu2023hallucidoctor}, for tree={fill=lightgray},text width=28em
                      ]
                      [
                        \textbf{LURE} \cite{zhou2023analyzing}, for tree={fill=lightgray},text width=28em
                      ]
                      [
                        \textbf{ VCD} \cite{leng2023mitigating}, for tree={fill=lightgray},text width=28em
                      ]
                    ]
                      [
                        \textbf{Fine-Tuning and Post-Processing}\\
                          [
                            \textbf{FDPO} \cite{gunjal2024detecting}, for tree={fill=lightgray},text width=28em
                          ]
                          [
                            \textbf{HA-DPO} \cite{zhao2023beyond}, for tree={fill=lightgray},text width=28em
                          ]
                          [
                            \textbf{InternLM-XComposer} \cite{zhang2023internlm}, for tree={fill=lightgray},text width=28em
                          ]
                      ]
                  [
                        \textbf{Generative Adversarial Networks}\\    
                      [
                        \textbf{VIGC} \cite{wang2024vigc}, for tree={fill=lightgray},text width=28em
                      ]
                      [
                        \textbf{ MARINE} \cite{zhao2024mitigating}, for tree={fill=lightgray},text width=28em
                      ]
                  ]
                  [
                    \textbf{Data-Centric Approaches}\\
                      [
                        \textbf{ViGoR} \cite{yan2024vigor}, for tree={fill=lightgray},text width=28em
                      ]
                      [
                        \textbf{Data Centric Approach} \cite{lu2024evaluation}, for tree={fill=lightgray},text width=28em
                      ]
                  ]
                  [
                    \textbf{Multi-Modal Fusion and Object-Level Modeling}\\
                      [
                        \textbf{MoF} \cite{tong2024eyes}, for tree={fill=lightgray},text width=28em
                      ]
                      [
                        \textbf{ObjMLM} \cite{dai2022plausible}, for tree={fill=lightgray},text width=28em
                      ]
                  ]
                  [
                    \textbf{Reinforcement Learning}\\
                      [
                        \textbf{Factually Augmented RLHF} \cite{sun2023aligning}, for tree={fill=lightgray},text width=28em
                      ]
                  ]
              ]
        ]
        [
            \textbf{Video} \S\ref{LVM1} , for tree={fill=bg25}, text width=18em
             [
                    \textbf{Detection}\\
                  [
                    \textbf{EMScore} \cite{shi2022emscore}, for tree={fill=pg68},text width=28em
                  ]
              ]
              [
                    \textbf{Mitigation}\\
                  [
                    \textbf{FactVC} \cite{liu2023models}, for tree={fill=pg60},text width=28em
                  ]
                  [
                    \textbf{CLearViD} \cite{chuang2023clearvid}, for tree={fill=pg60},text width=28em
                  ]
                  [
                    \textbf{MGAT} \cite{he2022multi}, for tree={fill=pg60},text width=28em
                  ]
              ]
        ]
        [
            \textbf{Audio} \S\ref{LAM} , for tree={fill=lime!50}, text width=18em
            [
                    \textbf{Detection}\\
                  [
                    \textbf{PAM} \cite{deshmukh2024pam}, for tree={fill=pg70},text width=28em
                  ]
                  [
                    \textbf{CompA} \cite{ghosh2023compa}, for tree={fill=pg70},text width=28em
                  ]
              ]
              [
                    \textbf{Mitigation}\\
                  [
                    \textbf{Latent Diffusion Models}\\
                      [
                        \textbf{MusicLDM} \cite{chen2024musicldm}, for tree={fill=red!50},text width=28em
                      ]
                      [
                        \textbf{Re-AudioLDM} \cite{yuan2024retrieval}, for tree={fill=red!50},text width=28em
                      ]
                  ]
                  [
                    \textbf{Retrieval-Based Approaches}\\
                      [
                        \textbf{RECAP} \cite{ghosh2024recap}, for tree={fill=red!50},text width=28em
                      ]
                  ]
                  [
                    \textbf{Self-Supervised and Contrastive Learning}\\
                      [
                        \textbf{SECap} \cite{xu2024secap}, for tree={fill=red!50},text width=28em
                      ]
                      [
                        \textbf{EnCLAP} \cite{kim2024enclap}, for tree={fill=red!50},text width=28em
                      ]
                  ]
                  [
                    \textbf{Cacophony} \cite{zhu2024cacophony}, for tree={fill=red!50},text width=28em
                  ]
              ]
        ]
      ]
    \end{forest}
 }
  \caption{Taxonomy of hallucination in large foundation models, organized around detection and mitigation techniques.}
  \label{taxo}
\end{figure*}

\begin{figure}[h]  
 \centering
\includegraphics[width=\linewidth]{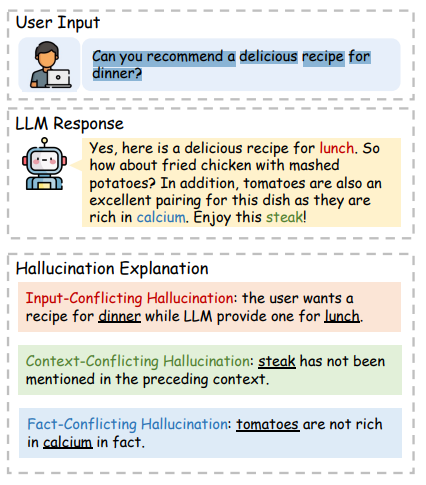}
 \caption{LLM responses showing the types of hallucinations, highlighted in \textcolor{red}{red}, \textcolor{teal}{green}, and \textcolor{blue}{blue} \cite{zhang2023siren}.}
\label{fig:llm-hallu}
\end{figure}

\section{Hallucination in Large Language Models}
\label{LLM}
Despite the progress of LLMs, a notable challenge persists in their proneness to hallucinate, impeding their practical implementation. For instance, the illustration in Figure~\ref{fig:llm-hallu} exemplifies the generated response by the LLM, showing indications of hallucination.




\subsection{Hallucination Detection and Mitigation}
Detecting hallucinations in LLMs is crucial for ensuring the credibility and reliability of their results, especially in scenarios requiring factual correctness. Existing fact-checking methods often rely on complex modules or external databases, requiring either output probability distributions or interfacing with external sources. The SelfCheckGPT method~\cite{manakul2023selfcheckgpt} offers a zero-resource black-box solution for detecting hallucinations in any LLM without relying on external resources. This method operates on the principle that an LLM familiar with a topic will produce consistent and comparable facts in its responses. In contrast, randomly sampled responses from an unfamiliar topic are likely to contain contradicting and hallucinated facts.
Continuing the exploration of methods for passage-level hallucination detection,~\citet{yang2023new} proposed a novel self-check approach based on reverse validation, aiming to automatically identify factual errors without external resources. They introduced a benchmark, Passage-level Hallucination Detection (PHD), generated using ChatGPT and annotated by human experts to assess different methods.
Assessing the accuracy of long text generated by LLMs is challenging because it often contains both accurate and inaccurate information, making simple quality judgments insufficient. To address this,~\citet{min2023factscore} introduced FACTSCORE (Factual Precision in Atomicity Score), a new evaluation method that breaks down text into individual facts and measures their reliability. 
~\citet{huang2023citation} introduced a unique strategy to mitigate hallucination risks in LLMs by drawing parallels with established web systems. They identified the absence of a "citation" mechanism in LLMs, which refers to acknowledging or referencing sources or evidence, as a significant gap. 

Addressing the need to identify factual inaccuracies in LLM-generated content,~\citet{rawte2024factoid} developed a multi-task learning (MTL) framework, integrating advanced long text embeddings like e5-mistral-7b-instruct, along with models such as GPT-3, SpanBERT, and RoFormer. This MTL approach demonstrated a 40\% average improvement in accuracy on the FACTOID benchmark compared to leading textual entailment methods. 
Hallucination mitigation efforts have predominantly relied on empirical methods, leaving uncertainty regarding the possibility of complete elimination. To tackle this challenge,~\citet{xu2024hallucination} introduced a formal framework defining hallucination as inconsistencies between computable LLMs and a ground truth function. The study examines existing hallucination mitigation strategies and their practical implications for real-world LLM deployment through this framework.~\citet{rawte2024sorry} introduced the Sorry, Come Again (SCA) prompting technique to address hallucination in contemporary LLMs. SCA enhances comprehension through optimal paraphrasing and injecting [PAUSE] tokens to delay LLM generation. It analyzes linguistic nuances in prompts and their impact on the hallucinated generation, emphasizing how prompts with lower readability, formality, or concreteness pose challenges.

\textbf{Benchmark Evaluation:}
In certain instances, LLMs engage in a phenomenon termed "hallucination snowballing," where they fabricate false claims to rationalize prior hallucinations despite acknowledging their inaccuracy~\cite{NEURIPS2023_81b83900}. To empirically explore this phenomenon,~\citet{zhang2023language} devised three question-answering datasets spanning diverse domains, wherein ChatGPT and GPT-4 often furnish inaccurate answers alongside explanations featuring at least one false claim. Significantly, the study suggests that the language model can discern these false claims as incorrect. Another benchmark dataset, FactCHD~\cite{chen2023unveiling}, was introduced to detect fact-conflicting hallucinations within intricate inferential contexts. It encompasses a range of datasets capturing different factuality patterns and integrates fact-based evidence chains to improve assessment accuracy.


\begin{figure} 
 \centering
\includegraphics[width=\linewidth]{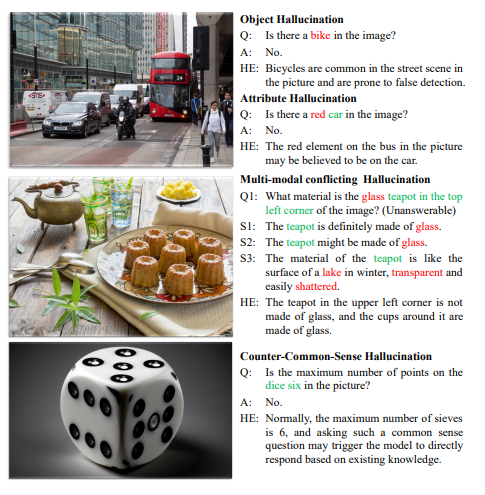}
 \caption{Four IVL-Hallu examples in Prompted Hallucination Dataset(PhD) ~\cite{liu2024phd} including visuals and the matching question-answer pairs and hallucination elements (HE). While words annotated in red do not exist or do not match within the image, words annotated in green have correspondences within the image. Question, Answer, and Statement are denoted by the letters Q, A, and S, respectively.}
\label{vlm}
\end{figure}

\section{Hallucination in Large Vision-Language Models}
\label{VLM1}
Large Vision-Language Models (LVLMs) have garnered significant attention in the AI community for their ability to handle visual and textual data simultaneously~\cite{ghosh2024exploring},~\cite{ghosh2024medsumm},~\cite{sahoo-etal-2024-enhancing},~\cite{ghosh2024sights},~\cite{ghosh2024clipsyntel}. Nonetheless, similar to LLMs, LVLMs also confront the issue of hallucination. Figure~\ref{vlm} illustrates an example of visual hallucination.

\subsection{Hallucination Detection and Mitigation}
~\citet{dai2022plausible} investigated the issue of object hallucinations in Vision-Language Pre-training (VLP) models, where textual descriptions generated by these models contain non-existent or inaccurate objects based on input images. 
~\citet{li2023evaluating} revealed widespread and severe object hallucination issues and suggested that visual instructions may influence hallucination, noting that objects frequently appearing in visual instructions or co-occurring with image objects are more likely to be hallucinated. To enhance the evaluation of object hallucination, they introduced a polling-based query method called POPE, which demonstrates improved stability and flexibility in assessing object hallucination. The absence of a standardized metric for assessing object hallucination has hindered progress in understanding and addressing this issue. To address this gap,~\citet{lovenia2023negative} introduced NOPE (Negative Object Presence Evaluation), a novel benchmark for evaluating object hallucination in vision-language (VL) models through visual question answering (VQA). Utilizing LLMs, the study generates 29.5k synthetic negative pronoun (NegP) data for NOPE. It extensively evaluates the performance of 10 VL models in discerning the absence of objects in visual questions, alongside their standard performance on visual questions across nine other VQA datasets. Most existing efforts focused primarily on object hallucination, overlooking the diverse types of LVLM hallucinations.~\citet{liu2024phd} delved into Intrinsic Vision-Language Hallucination (IVL-Hallu) and proposed several novel IVL-Hallu tasks categorized into four types: attribute, object, multi-modal conflicting, and counter-common-sense hallucination. To assess and explore IVL-Hallu, they introduced a challenging benchmark dataset and conducted experiments on five LVLMs, revealing their incapacity to effectively address the proposed IVL-Hallu tasks. To mitigate object hallucination in LVLMs without resorting to costly training or API reliance,~\citet{zhao2024mitigating} introduced MARINE, which is both training-free and API-free. MARINE enhances the visual understanding of LVLMs by integrating existing open-source vision models and utilizing guidance without classifiers to integrate object grounding features, thereby improving the precision of the generated outputs. Evaluations across six LVLMs reveal MARINE's effectiveness in reducing hallucinations and enhancing output detail, validated through assessments using GPT-4V.~\cite{deng2024seeing} introduced a CLIP-Guided Decoding (CGD) training-free approach to reduce object hallucination at decoding time.



HalluciDoctor~\cite{yu2023hallucidoctor} tackled hallucinations in Multi-modal Large Language Models (MLLMs) by using human error detection to identify and eliminate various types of hallucinations. By rebalancing data distribution via counterfactual visual instruction expansion, they successfully mitigate 44.6\% of hallucinations while maintaining competitive performance. Despite proficiency in visual semantic comprehension and meme humor, MLLMs struggle with chart analysis and understanding. Addressing this,~\citet{xu2023chartbench} proposed ChartBench, a benchmark assessing chart comprehension. ChartBench exposes MLLMs' limited reasoning with complex charts, prompting the need for novel evaluation metrics like Acc+ and a handcrafted prompt, ChartCoT.~\citet{zhang2023internlm} introduced InternLM-XComposer, an LVLM aimed at designed to address the challenge of hallucination in image-text comprehension and composition. The performance of InternLM-XComposer's text-image composition is evaluated through a robust procedure involving both human assessment and comparison to GPT4-Vision, with the model demonstrating competitive performance against solutions like GPT4-V and GPT3.5.~\citet{wang2024mitigating} proposed the Instruction Contrastive Decoding (ICD) method to reduce hallucinations during LVLM inference. 
A recent study~\cite{Huang_2024_CVPR} proposed a novel decoding approach that introduces an over-confidence penalty and a retrospective allocation strategy to mitigate hallucination issues without requiring additional data or retraining.

\subsection{Benchmark Evaluation}
The current methods of developing LVLMs rely heavily on annotated benchmark datasets, which can exhibit domain bias and limit model generative capabilities. To address this,~\citet{li2023stablellava} proposed a novel data collection approach that synthesizes images and dialogues synchronously for visual instruction tuning, yielding a large dataset of image-dialogue pairs and multi-image instances. 
~\citet{huang2024visual} introduced VHTest, a benchmark dataset with 1,200 diverse visual hallucinations (VH) instances across 8 VH modes. Evaluation of three SOTA MLLMs showed varying performance, with GPT-4V exhibiting lower hallucination than MiniGPT-v2. 
~\citet{rawte2024visual} categorized visual hallucination in VLMs into eight orientations and introduced a dataset of 2,000 samples covering these types. They proposed three main categories of methods to mitigate hallucination: data-driven approaches, training adjustments, and post-processing techniques.~\citet{wang2024vigc} proposed the Visual Instruction Generation and Correction (VIGC) framework to address the scarcity of high-quality instruction-tuning data for MLLMs. VIGC enables MLLMs to generate diverse instruction-tuning data while iteratively refining its quality through Visual Instruction Correction (VIC), mitigating hallucination risks. The framework produces diverse, high-quality data for fine-tuning models, validated through evaluations, improving benchmark performance, and overcoming language-only data limitations.


\begin{figure*} 
 \centering
\includegraphics[width=0.9\linewidth]{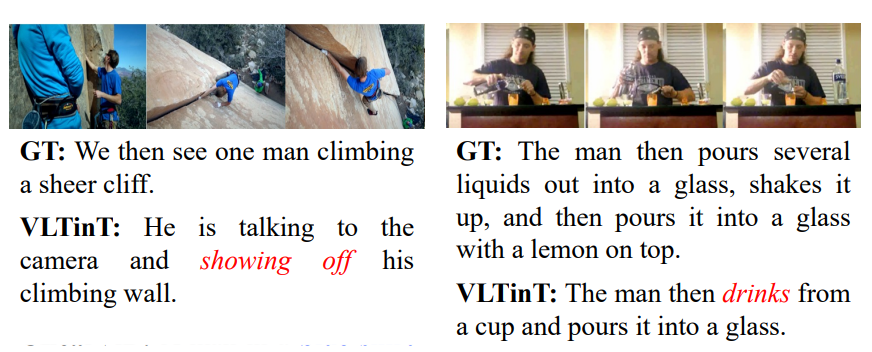}
 \caption{A video featuring descriptions generated by VLTinT model and ground truth (GT) with description errors highlighted in red italics. \cite{chuang2023clearvid}.}
\label{fig:Video}
\end{figure*}

\section{Hallucinations in Large Video Models}
\label{LVM1}
Large video models (LVMs) represent a significant advancement, allowing for processing video data at scale. Despite their potential for various applications like video understanding and generation, LVMs face challenges with hallucinations, where misinterpretations of video frames can result in artificial or inaccurate visual data. 
Figure~\ref{fig:Video} demonstrates the instances of hallucination observed in LVMs.

\subsection{Hallucination Detection and Mitigation}
The intricate task of dense video captioning, involving the creation of descriptions for multiple events within a continuous video, necessitates a thorough understanding of video content and contextual reasoning to ensure accurate description generation. However, this endeavor faces numerous challenges, potentially resulting in instances of inaccuracies and hallucinations~\cite{iashin2020multi},~\cite{suin2020efficient}. Traditional methods detect event proposals first, then caption subsets, risking hallucinations due to overlooking temporal dependencies. To address this,~\citet{mun2019streamlined} introduced a novel approach to modeling temporal dependencies and leveraging context for coherent storytelling. By integrating an event sequence generation network and a sequential video captioning network trained with reinforcement learning and two-level rewards, the model captures contextual information more effectively, yielding coherent and accurate captions while minimizing the risk of hallucinations.~\citet{liu2023models} introduced a novel weakly-supervised, model-based factuality metric called FactVC, which outperforms previous metrics. Furthermore, they provided two annotated datasets to promote further research in assessing the factuality of video captions.~\citet{wu2023context} proposed a context-aware model that incorporates information from past and future events to influence the description of the current event conditionally. Their approach utilizes a robust pre-trained context encoder to encode information about the surrounding context events, which is then integrated into the captioning module using a gate-attention mechanism. Experimental findings on the YouCookII and ActivityNet datasets demonstrate that the proposed context-aware model outperforms existing context-aware and pre-trained models by a significant margin. To enhance dense video captioning, ~\citet{zhou2024streaming} introduced a streaming model comprising a memory module for long video handling and a streaming decoding algorithm enabling predictions before video completion. This approach notably boosts performance on dense video captioning benchmarks such as ActivityNet, YouCook2, and ViTT.

Video infilling and prediction tasks are crucial for assessing a model's ability to comprehend and anticipate the temporal dynamics within video sequences~\cite{hoppe2022diffusion}. To address this,~\citet{himakunthala2023let} introduced an inference-time challenge dataset containing keyframes with dense captions and structured scene descriptions. This dataset contains keyframes supplemented with unstructured dense captions and structured FAMOUS: \textit{(Focus, Action, Mood, Objects, and Setting)} scene descriptions, providing valuable contextual information to support the models' understanding of the video content. They employed various language models like GPT-3, GPT-4, and Vicuna with greedy decoding to mitigate hallucination risks.
Prominent developments in video inpainting have been observed recently, especially in situations where explicit guidance like optical flow helps to propagate missing pixels across frames~\cite{ouyang2021internal}. However, difficulties and constraints occur from a lack of cross-frame information.~\citet{yu2023deficiency} aimed to tackle the opposite issue rather than depending on using pixels from other frames. The suggested method presents a Deficiency-aware Masked Transformer (DMT), a dual-modality-compatible inpainting framework. This approach improves handling scenarios with incomplete information by pre-training an image inpainting model to serve as a prior for training the video model.

Understanding scene affordances, which involve potential actions and interactions within a scene, is crucial for comprehending images and videos.~\citet{kulal2023putting} introduced a method for realistically inserting people into scenes. The model seamlessly integrates individuals into scenes by deducing realistic poses based on the context and ensuring visually pleasing compositions.
~\citet{chuang2023clearvid} introduced CLearViD, a transformer-based model that utilizes curriculum learning techniques to enhance performance. By adopting this approach, the model acquires more robust and generalizable features. Furthermore, CLearViD incorporates the Mish activation function to address issues like vanishing gradients, thereby reducing the risk of hallucinations by introducing nonlinearity and non-monotonicity. Extensive experiments and ablation studies validate the effectiveness of CLearViD, with evaluations on ActivityNet captions and YouCook2 datasets showcasing significant improvements over existing SOTA models in terms of diversity metrics. 

\subsection{Benchmark Evaluation}
~\citet{zhang2006video} created a novel two-level hierarchical fusion method to hallucinate facial expression sequences from training video samples using only one frontal face image with a neutral expression. To effectively train the system, they introduced a dataset specifically designed for facial expression hallucination, which included 112 video sequences covering four types of facial expressions (happy, angry, surprise, and fear) from 28 individuals, resulting in the generation of reasonable facial expression sequences in both the temporal and spatial domains with less artifact. In the realm of video understanding, the development of end-to-end chat-centric systems has become a growing area of interest.~\citet{zhou2018towards} assembled the YouCook2 dataset, an extensive set of cooking videos with temporally localized and described procedural segments, to facilitate procedure learning tasks.~\citet{li2023videochat} introduced "VideoChat," a novel approach integrating video foundation models and LLMs through a learnable neural interface to enhance spatiotemporal reasoning, event localization, and causal relationship inference in video understanding. The researchers constructed a video-centric instruction dataset with detailed descriptions and conversations, emphasizing spatiotemporal reasoning and causal relationships. To counteract model hallucination, they employed a multi-step process to condense video descriptions into coherent narratives using GPT-4 and refined them to improve clarity and coherence. To explore the challenge of deducing scene affordances,~\citet{kulal2023putting} curated a dataset of 2.4M video clips, showcasing a variety of plausible poses that align with the scene context.

\begin{figure*}
 \centering
\includegraphics[width=\linewidth]{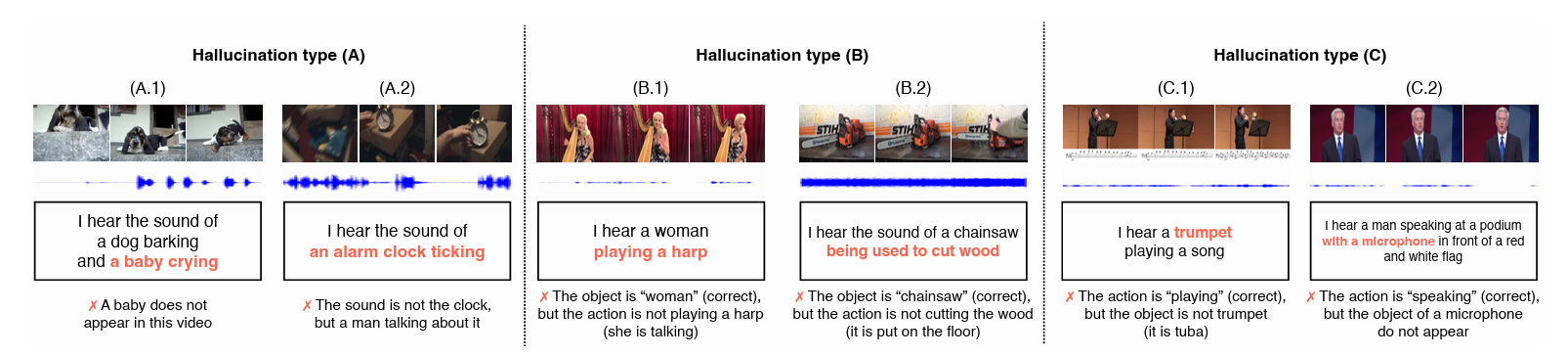}
 \caption{ Audio hallucination examples for each classes - Type A: \textit{Involving hallucinations of both objects and actions:} Type B: \textit{Featuring accurate objects but hallucinated actions;} Type C: \textit{Displaying correct actions but hallucinated objects} \cite{nishimura2024audio}.}
\label{fig:Audio hallucination}
\end{figure*}

\section{Hallucinations in Large Audio Models}
\label{LAM}
Large audio models (LAMs) have emerged as a powerful tool in the realm of audio processing and generation, with a wide range of applications like speech recognition, music analysis, audio synthesis, and captioning~\cite{latif2023sparks},
~\cite{ghosal2023text}.
While demonstrating remarkable capabilities, LAMs are susceptible to hallucinations – anomalies ranging from generating unrealistic audio by combining fabricated snippets to injecting false information like quotes or facts into summaries. Additionally, they may fail to accurately capture the inherent features of audio signals, such as timbre, pitch, or background noise~\cite{shen2023naturalspeech}. Figure~\ref{fig:Audio hallucination} presents one example of audio hallucinations.

\subsection{Hallucination Detection and Mitigation}
In the realm of audio captioning, where natural language descriptions for audio clips are automatically generated, a significant challenge arises from the over-reliance on the visual modality during the pre-training of audio-text models. This reliance introduces data noise and hallucinations, ultimately undermining the accuracy of the resulting captions. To address this issue,~\citet{xu2023blat} introduced an AudioSet tag-guided model designed to bootstrap large-scale audio-text data (BLAT). Notably, this model sidesteps the incorporation of video, thus minimizing noise associated with the visual modality. The experimental findings across a range of tasks, including retrieval, generation, and classification, validate the effectiveness of BLAT in mitigating hallucination issues.

Speech emotions play a crucial role in human communication and find extensive applications in areas such as speech synthesis and natural language understanding. However, traditional categorization approaches may fall short of capturing the nuanced and intricate nature of emotions conveyed in human speech~\cite{jiang2019parallelized},~\cite{han2021automated},~\cite{ye2021improving}.
SECap~\cite{xu2024secap}, a framework designed for speech emotion captioning. It aims to capture the intricate emotional nuances of speech using natural language. SECap utilizes various components, including LLaMA as the text decoder, HuBERT as the audio encoder, and Q-Former as the Bridge-Net, to generate coherent emotion captions based on speech features. Audio-language models, despite their capability for zero-shot inference, confront challenges like hallucinating task-specific details despite strong performance. To address this,~\citet{elizalde2024natural} introduced the Contrastive Language-Audio Pretraining (CLAP) model. Pre-trained with 4.6 million diverse audio-text pairs, CLAP features a dual-encoder architecture, enhancing representation learning for improved task generalization across sound, music, and speech domains.


\subsection{Benchmark Evaluation}
To address the scarcity of data in the specific domain of music captioning,~\citet{doh2023lp} introduced LP-MusicCaps, a comprehensive dataset comprising 0.5 million audio clips accompanied by approximately 2.2 million captions. Leveraging LLMs, they trained a transformer-based music captioning model with the dataset and assessed its performance under zero-shot and transfer-learning scenarios, demonstrating its superiority over supervised baseline models.~\citet{nishimura2024audio} investigated audio hallucinations in large audio-video language models, where audio descriptions are generated primarily based on visual information, neglecting audio content. They have classified these hallucinations into three distinct types such as \textit{Involving hallucinations of both objects and actions}, \textit{Featuring accurate objects but hallucinated actions}, and \textit{Displaying correct actions but hallucinated objects} as represented in Fig.~\ref{fig:Audio hallucination}. They gathered 1000 sentences by soliciting audio information and then annotated them to determine whether they contained auditory hallucinations. To assess compositional reasoning in LAMs,~\citet{ghosh2023compa} introduced CompA, consisting of two expert-annotated benchmarks and employed to fine-tune CompA-CLAP with a novel learning approach, leading to improved compositional reasoning abilities compared to baseline models on related tasks.

\section{Future Directions}
\label{FD}
Researchers are actively investigating techniques to mitigate hallucinations, which is crucial for sensitive applications~\cite{tonmoy2024comprehensive},~\cite{rawte2023survey}. The Key future directions can include:

\textbf{Data Resources:} Fine-tuning carefully curated high-quality data, integrating structured knowledge from knowledge graphs, and employing task/domain-specific alignment techniques to enhance accuracy and relevance.

\textbf{Automated Evaluation:} Developing specialized metrics for factual accuracy and coherence, combining automated evaluation with human judgments, adversarial testing to identify weaknesses, and fine-tuning fact-checking datasets.

\textbf{Improving Detection and Mitigation:} Leveraging reasoning mechanisms (e.g., Chain of Thought~\cite{wei2022chain}, Tree of Thought~\cite{yao2024tree}), knowledge graph integration, specialized fact-checking models, bias mitigation techniques, active learning methodologies, and ethical guidelines/regulatory frameworks.

\textbf{Multimodal Hallucination:} Data-centric initiatives, cross-modal alignment, architectural innovations, standardized benchmarking, reframing hallucination as a feature, and enhancing interpretability and trust for reliable multimodal AI systems.

\section{Conclusion}
This survey paper systematically categorizes existing research on hallucination within FMs, providing comprehensive insights into critical aspects such as detection, mitigation, tasks, datasets, and evaluation metrics. It addresses the pressing issue of hallucination in FMs, acknowledging its widespread impact across various domains. The paper underscores the importance of addressing this challenge by examining recent advancements in detection and mitigation techniques, given FMs' indispensable role in critical tasks. Its primary contribution is presenting a structured taxonomy for classifying hallucination in FMs, spanning text, image, video, and audio domains.

\section{Limitation}
Previous survey papers primarily focused on hallucination in large language models and did not extensively cover hallucinations in vision, audio, and video modalities. This survey paper aims to provide a comprehensive overview of hallucinations across all modalities, considering that hallucinations can occur in any large foundation model. Despite our efforts to provide a comprehensive summary of recent advancements related to hallucination techniques in all foundational models, we acknowledge that we may miss some relevant work in the field and have covered the papers till May 2024. 

\label{ex}
\bibliography{anthology,custom}
\bibliographystyle{acl_natbib}
\newpage
\section{Appendix}
\subsection{Table}
\label{ta}
We have provided a comprehensive summary of the methodologies pertaining to hallucination techniques in large foundational models in Table~\ref{R1}, detailing their approaches to hallucination detection, mitigation, task considerations, datasets utilized, and evaluation metrics employed. This will offer readers a concise overview of recent advancements in this field. 


\begin{table*}[hbt!]
    \centering
    \scalebox{0.573}{
    \begin{tabular}{|c |c c c c c c |}
    \hline
     \multirow{30}{*}{\rotatebox[origin=c]{90}{TEXT}} & \textbf{Paper} & \textbf{Detection} & \textbf{Mitigation} & \textbf{Task}  & \textbf{Dataset(s)}  & \textbf{Evaluation Metric(s)}   \\ \cline{2-7}
        & \cite{manakul2023selfcheckgpt}  &  Yes & No & QA & Wikibio & Entropy  \\ 
        & \cite{li2022valhalla} & Yes & Yes & QA, Dialog summarization &  Halueval & Automatic \\
        & \cite{mundlerself} & Yes & Yes & Text generation & Manual & F1 Score\\
        & \cite{chen2023purr} & No & Yes & Editing for attribution & MCQ, Dialog & Attribution, Preservation \\
        & \cite{zhang2023mitigating} & No & Yes & Question knowledge alignment & Fuzzy QA & Attributable to Identified Sources \\
        & \cite{zhang2023language} & Yes & No & QA & Manual & Accuracy \\
        & \cite{peng2023check} & No & Yes & Task-oriented dialog &  News, Customer service & F1 Score, Bleu-4  \\
        & \cite{cui2023chatlaw} &  No & Yes & QA & Manual & Ranking \\
        & \cite{azaria2023internal}  & Yes & No & Classification & Manual & Accuracy \\
        & \cite{li2023chain} & Yes & Yes & Knowledge-intensive tasks & Fever, QA & Accuracy \\
        & \cite{elaraby2023halo} & Yes & Yes & Consistency, Actuality, QA & Manual NBA domain & Pearson Correlation Coefficient \\
        & \cite{varshneystitch} & Yes & Yes & Text generation & Wikibio & Percentage of mitigated hallucination \\
        & \cite{jha2023dehallucinating} & Yes & No & Dialog & N/A & N/A \\
        & \cite{pal2023medhalt} & No & No & Reasoning hallucination & Med-Halt & Accuracy, Pointwise Score \\
        & \cite{mckenna2023sources} & Yes & No & Textual entailment & Altered Directional Inference & Entailment Probability \\
        & \cite{guerreiro2023hallucinations} & Yes & Yes & MT & FLores 101, WMT ,TICO & BLEU \\
        & \cite{huang2023citation} &  Yes & Yes & N/A & N/A & N/A \\
        & \cite{luo2023zero} & Yes & Yes & Concept extraction & Concept-7 & AUC, Accuracy, F1 Score \\
        & \cite{gao2022rarr} & Yes & Yes & Editing attribution & NQ, SQA & Auto-AIS (Attr\_{auto}) \\
        & \cite{yang2023new} & Yes & No & \begin{tabular}[c]{@{}c@{}}Detect factual\\ errors automatically\end{tabular} & PHD, WikiBio-GPT3 &\begin{tabular}[c]{@{}c@{}} Precision, Recall, \\F1 Score, Accuracy\end{tabular} \\
        & \cite{min2023factscore} & Yes & Yes & Fact verification & Manual(Wikipedia) & FActScore \\
        & \cite{rawte2024factoid} & Yes & Yes & Factual inaccuracies detection & FACTOID & HV I\_{auto} \\
        & \cite{ahmad2023creating} & Yes & Yes & Hallucination in healthcare & N/A & FActScores \\
        & \cite{ji2023towards} & Yes & Yes & Generative and knowledge-intensive & \begin{tabular}[c]{@{}c@{}}PubMedQA, MEDIQA2019,\\  MedQuAD, and MASH-QA\end{tabular} &  \begin{tabular}[c]{@{}c@{}} Unigram F1, ROUGE-L,\\  Med-NLI, and CTRLEval\end{tabular}  \\
        & \cite{kang2023deficiency} & Yes & Yes & Hallucination in finance & N/A & FActScores \\
        & \cite{roychowdhury2024journey} & No & Yes & QA & N/A & N/A \\
        & \cite{savelka2023explaining} & No & Yes & Factual evaluation in legislation & N/A & N/A \\
        & \cite{dahl2024large} & Yes & No & Legal hallucination & Manual & N/A \\ 
        & \cite{rawte2024sorry}  &  No & Yes & Comprehension enhancement & SCA-90K & Cosine similarity  \\ 
    \hline

    \multirow{28}{*}{\rotatebox[origin=c]{90}{IMAGE}} & \cite{li2023evaluating} & Yes & No & Evaluation of object hallucination & MSCOCO & CHAIR, POPE \\
    & \cite{gunjal2024detecting} & Yes & Yes & VQA & M-Hall Detect & Accuracy \\
    & \cite{dai2022plausible} & No & Yes & Image captioning & CHAIR & CIDEr \\
    & \cite{lovenia2023negative} & Yes & No & Object hallucination & NOPE & \begin{tabular}[c]{@{}c@{}}METEOR, Exact match accuracy,\\ NegP Accuracy\end{tabular}  \\
    & \cite{liu2024phd} & Yes & No & Intrinsic vision-language hallucination & PhD & Accuracy \\
    & \cite{zhao2024mitigating} & Yes & Yes &  Non-existing object hallucination & MSCOCO & CHAIR, POPE, GPT-4V, recall \\
    & \cite{huang2024visual} & Yes & No & Visual hallucination & YNQ, OEQ & Accuracy \\
    & \cite{rawte2024visual} & Yes & No & Video captioning &  ActivityNet-Fact, YouCook2-Fact & FactVC \\
    & \cite{wang2024vigc} &  No & Yes & \begin{tabular}[c]{@{}c@{}}Generate instruction data\\  for vision-language\end{tabular} & \begin{tabular}[c]{@{}c@{}} VIGC-LLaVA-COCO,\\ VIGC-LLaVA-Objects365\end{tabular} & Conv, Detail, Complex \\ 
    & \cite{yu2023hallucidoctor} & Yes & Yes & \begin{tabular}[c]{@{}c@{}} Machine-generated \\visual instruction\end{tabular} & LLaVA-Instruction-158K  & CHAIR \\
    & \cite{guan2023hallusionbench} & No & Yes & Visual questions & HallusionBench & Accuracy \\
    & \cite{liu2023mitigating} & Yes & Yes & Vision language & LRV-Instruction & GAVIE \\
    & \cite{xu2023chartbench} & Yes & No & \begin{tabular}[c]{@{}c@{}}Evaluation of MLLMs on\\ chart comprehension\end{tabular} & ChartBench & Acc+  \\
    & \cite{lu2024evaluation} &Yes & Yes & Vision language & MSG-MCQ & Accuracy \\
    & \cite{tong2024eyes} & Yes & No & Visual question answering & MMVP, VQA & Accuracy \\
    & \cite{liao2023revo} & Yes & No & Vision language & REVO-LION & Meta Quality (MQ) \\
    & \cite{hu2023ciem} & Yes & Yes & \begin{tabular}[c]{@{}c@{}}Visual captioning, \\Visual question answering\end{tabular} & CIEM & \begin{tabular}[c]{@{}c@{}}Accuracy, Precision,\\ Recall, F1 Score\end{tabular} \\
    & \cite{jing2023faithscore} & Yes & No & Meta-evaluation & LLaVA-1k, MSCOCO-Cap & FAITHSCORE   \\
    & \cite{changpinyo2022maxm} & No & Yes & Multilingual visual question answering & MaXM &  Accuracy \\
    & \cite{wang2023evaluation} & Yes & No & Content generation & N/A & Precision, Recall, F1 Score  \\
    & \cite{sun2023aligning} & No & Yes & Visual-language alignment & MMHAL-BENCH &  N/A \\
    & \cite{bai2023touchstone} & Yes & No & \begin{tabular}[c]{@{}c@{}}Evaluate hallucination of\\ vision language model\end{tabular} & TouchStone & Hallucination Score  \\
    & \cite{zhou2023analyzing} & No & Yes  & Hallucination mitigation in LVMs & MSCOCO & CHAIR, BLEU, CLIP  \\
    & \cite{yan2024vigor} & No & Yes & Visual grounding & MMViG & \begin{tabular}[c]{@{}c@{}} HL, CA, AA, \\RA, RL, RS, DL\end{tabular}\\
    & \cite{zhao2023beyond} & Yes & Yes &  Overcome hallucination in LVMs & POPE, SHR & Accuracy, Precision, F1 Score  \\
    & \cite{zhang2023internlm} & No & Yes & \begin{tabular}[c]{@{}c@{}}Image text comprehension\\ and composition\end{tabular} & \begin{tabular}[c]{@{}c@{}}MMBench, SeedBench, QBench, \\MMBench-CN, Chinese Bench\end{tabular} & \begin{tabular}[c]{@{}c@{}}LR, AR, RR,\\ FP-C, FP-S, CP\end{tabular}  \\

    \hline
    \multirow{7}{*}{\rotatebox[origin=c]{90}{VIDEO}} & \cite{kulal2023putting} &  No & Yes &  Affordance prediction & Manual &  FID, FCKh \\
    & \cite{himakunthala2023let} & No & Yes & Video infilling, Scene prediction & Manual & N/A \\
    & \cite{li2023videochat} & No & Yes & Visual dialogue & Manual & N/A \\
    & \cite{zhou2024streaming} & No & Yes & Video captioning & \begin{tabular}[c]{@{}c@{}}ActivityNet Captions,\\ YouCook2, ViTT \end{tabular}& CIDER, METEOR, SODAc \\
    & \cite{hoppe2022diffusion} & Yes & No & Video prediction & BAIR, Kinetics 600, UCF-101 & Frechet Video Distance \\
    & \cite{chuang2023clearvid} & No & Yes & Video description &  Activity Net Captions, YouCook2 & \begin{tabular}[c]{@{}c@{}}METEOR, ROUGE\_L, CIDER, \\BLEU\_4, DIV-2, RE\_4 \end{tabular} \\
    
    \hline
    \multirow{5}{*}{\rotatebox[origin=c]{90}{AUDIO}} & \cite{li2023audio} & No & Yes & Classification & Manual &  Mean avg precision \\
    & \cite{doh2023lp} & No  & Yes   &  Audio captioning  & LP MusicCaps &  BLEU \\
    & \cite{xu2023blat} & No & Yes & Caption generation & AudioCaps & R@K, COCO \& FENCE \\
    & \cite{liu2023models} & No & Yes & Audio captioning & MusciCaps & BLEU\\
    & \cite{nishimura2024audio} & Yes & No & Evaluation of LAMs & LAION\_CLAP,MS\_CLAP & Recall, Precision, F1 Score \\ 
    \hline
      
      \hline 
    \end{tabular}}
    \caption{Overview of the hallucination detection and mitigation landscape in FMs across modalities (Text, Image, Video, and Audio). Each work is categorized based on factors such as detection, mitigation, tasks, datasets, and evaluation metrics.}
    \label{R1}
    
\end{table*}

\end{document}